\documentclass[referee,pdflatex,sn-vancouver-num]{sn-jnl}

\usepackage{graphicx}%
\usepackage{multirow}%
\usepackage{amsmath,amssymb,amsfonts}%
\usepackage{amsthm}%
\usepackage{mathrsfs}%
\usepackage[title]{appendix}%
\usepackage{xcolor}%
\usepackage{textcomp}%
\usepackage{manyfoot}%
\usepackage{booktabs}%
\usepackage{algorithm}%
\usepackage{algorithmicx}%
\usepackage{algpseudocode}%
\usepackage{listings}%

\usepackage{hyperref}
\usepackage{soul}
\setcitestyle{super,comma,open={},close={}} 

\usepackage{tikz}
\usetikzlibrary{shapes.geometric, arrows.meta, positioning, calc, fit, backgrounds}

\theoremstyle{thmstyleone}%
\theoremstyle{thmstyletwo}%

\theoremstyle{thmstylethree}%

\begin{document}


\title[Article Title]{Climate Prompting: Generating the Madden-Julian Oscillation using Video Diffusion and Low-Dimensional Conditioning}

\author*[1]{\fnm{Sulian} \sur{Thual}}\email{sthual@clemson.edu}
\author[1]{\fnm{Feiyang} \sur{Cai}}
\author[1]{\fnm{Jingjing} \sur{Wang}}
\author*[1]{\fnm{Feng} \sur{Luo}}\email{luofeng@clemson.edu}
\affil[1]{\orgname{School of Computing}, \orgaddress{\street{100 McAdams Hall}, \city{Clemson}, \state{SC}, \country{USA}}}

\abstract{
%
Generative Deep Learning is a powerful tool for modeling of the Madden–Julian oscillation (MJO) in the tropics, yet its relationship to traditional theoretical frameworks remains poorly understood.
Here we propose a video diffusion model, trained on atmospheric reanalysis, to synthetize long MJO sequences conditioned on key low-dimensional metrics. 
The generated MJOs capture key features including composites, power spectra and multiscale structures including convectively coupled waves, despite some bias.
We then ``prompt" the model to generate more tractable MJOs based on intentionally idealized low-dimensional conditionings, for example a perpetual MJO, an isolated modulation by seasons and/or the El Niño–Southern Oscillation, and so on. This enables deconstructing the underlying processes and identifying physical drivers.
The present approach provides a practical framework for bridging the gap between low-dimensional MJO theory and high-resolution atmospheric complexity and will help tropical atmosphere prediction.
} 


\maketitle

\section*{Introduction}\label{sec_intro}


The Madden–Julian oscillation (MJO) is the dominant compononent of intraseasonal variability in the tropics \cite{zhang2005madden}. It manifests as an equatorial, planetary-scale wave that originates in the Indian Ocean and propagates eastward across the western Pacific at approximately \(5\text{\ m.\ s}^{-1}\), as well as a prominent signal around zonal wavenumbers 1-3 and frequencies 40-90 days\cite{wheelerkiladis1999}. The MJO modulates monsoon evolution, mid-latitude predictability, and the onset of El Niño events, making it a critical factor in daily to seasonal forecasting.
However, the MJO is inherently intermittent and disorganized; its multiscale nature, driven by the non-linear coupling of convection and large-scale circulation, remains a primary source of forecast uncertainty. This complexity leaves a persistent gap in our ability to simulate and predict tropical weather transitions, challenging both current theoretical frameworks and numerical models\cite{zhang2020four, delaunaychristensen2022mjodl}.

Deep learning has rapidly transformed weather and climate science, enhancing both predictive skill and dynamical understanding\cite{pathak2022fourcastnet, bi2023panguweatherera5, lam2023graphcastera5}. These advances extend to the intraseasonal scale, where diffusion models have recently emerged as a primary focus for generative applications\cite{chen2023fuxi, stock2024diffobs, ren2025videodiffcyclones}. Conversely, traditional MJO theory utilizes low-order models to capture fundamental features such as the MJO's characteristic eastward propagation, quadrupole structure, intermittency, and so on \cite{zhang2020four, majda2009skeleton, thual2014stochastic}. However, a significant dimensionality gap remains between these two approaches: deep learning excels at capturing high-dimensional complexity but often lacks transparency, while low-order models offer physical tractability at the expense of spatial resolution. 
This fundamental disconnect persists despite diverse efforts to reconcile data-driven power with physical rigor, ranging from the use of Explainable AI (XAI) to probe internal model representations\cite{martin2022xai, shin2024deep} to the development of physics-informed constraints\cite{kashinath2021physics} and interpretable frameworks that identify low-dimensional convective manifolds\cite{behrens2022non, yao2024machine}. Such efforts underscore an urgent need for new architectures that align high-capacity generative manifolds with the reduced-order drivers central to our dynamical understanding.
 
Here we bridge the dimensionality gap between theoretical indices and high-resolution atmospheric fields by introducing a conditional video diffusion framework for MJO synthesis. 
 We propose ``low-dimensional climate prompting," a generative paradigm that treats fundamental physical drivers—including a modified Real-time Multivariate MJO (RMM) index\cite{wheeler2004rmm, stachnik2015evaluating}, seasonal sinusoidal embeddings, and ENSO states—as conditioning tokens. 
 Unlike traditional methods that use these indices for deterministic forecasting\cite{pathak2022fourcastnet} or linear analog reconstruction, our model leverages the probabilistic nature of diffusion to map low-order states onto a manifold of physically consistent, high-dimensional realizations. We demonstrate the utility of this promptable interface by synthesizing prolonged MJO sequences that capture the core characteristics of the observed oscillation.

By intentionally generating semi-realistic, idealized or even counterfactual\cite{kamphuis2011retrograde} MJO evolutions, we show how the present approach can be used to deconstruct some of the underlying MJO physical mechanisms, offering a powerful tool for hypothesis testing in tropical dynamics. This establishes a bidirectional link between generative AI and dynamical theory, effectively transforming key climate metrics into tunable generative drivers. Thus the present approach provides a practical framework for integrating deep learning with physical interpretability in Earth system modeling.


\section*{Model Validation}\label{sec_validation}

\subsection*{Training Record}

The present video diffusion model uses a standard U-Net architecture, is trained on the ECMWF Reanalysis v5 (ERA5), and uses Brick-Wall Denoising for extended sampling, as documented in the methods section\cite{ho2020ddpm, hersbach2020era5, yuan2025brickwall}. Ultimately the model generates long MJO sequences (e.g. 60 years) as prompted from low-dimensional climate metrics (i.e. conditionings), as listed in Tab.~\ref{tab:conditionings}. 

\begin{figure*}[h!]
\doublespacing 
\centering
\includegraphics[width=1\linewidth]{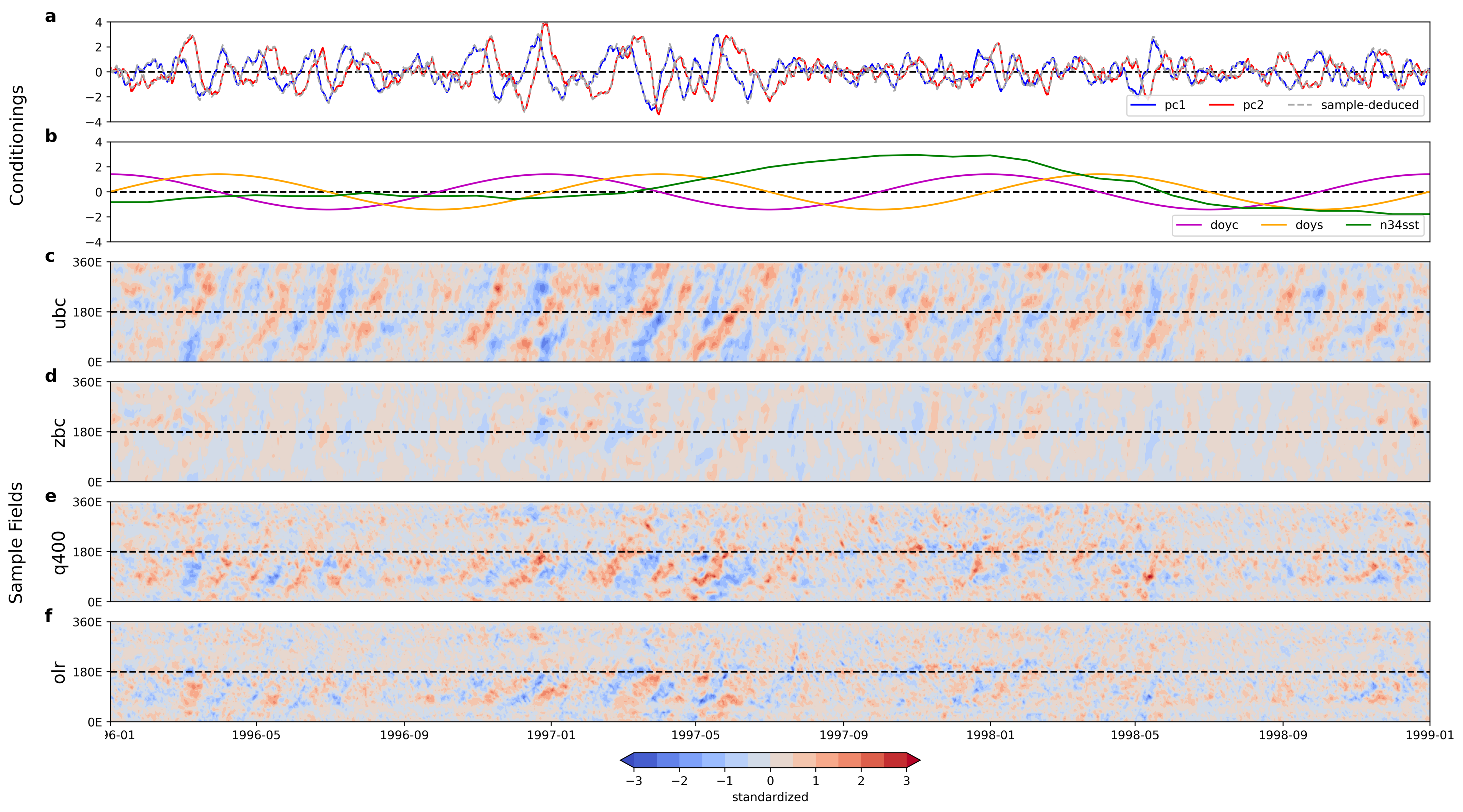}
\caption{Model generated MJO sequence. 
\textbf{a}, Conditionings pc1 and pc2, as a function of time. 
\textbf{b}, Conditionings doys, doys and n34sst. 
\textbf{c,d,e,f} Hovmollers of sample fields, as a function of time and longitude: zonal winds (UBC), geopotential (ZBC), moisture (Q400) and convection (OLR). Fields are averaged over equatorial band (15N-15S). 
Here the model is prompted with observed conditionings from ERA5 1960-2022 (thus it regenerates the training record). All conditionings and fields are standardized. The principal components deduced directly from the sample fields match the conditionings (gray dashed lines in a).
}\label{fig:fig_hovs_model_training}
\end{figure*}

Here we validate the video diffusion model by regenerating the training record, which details the generative process and highlights some limitations such as certain biases MJO in representation. In other words, we sample a long video 
from a prompt conditioning that is exactly as in the ERA5 observations 1960-2022. This is illustrated in Fig.~\ref{fig:fig_hovs_model_training}. The model replicates intermittent MJO sequences as in the original record, which amplitude and phase closely follow the conditionings pc1, pc2 (Tab.~\ref{tab:conditionings}). In fact, the principal components deduced from the sampled video closely match the conditionings (see methods). The other conditionings (doyc, doys and n34sst, Tab.~\ref{tab:conditionings}) instead modulate the MJO characteristics, as discussed hereafter.  

\begin{figure*}[h!]
\doublespacing 
\centering
\includegraphics[width=1\linewidth]{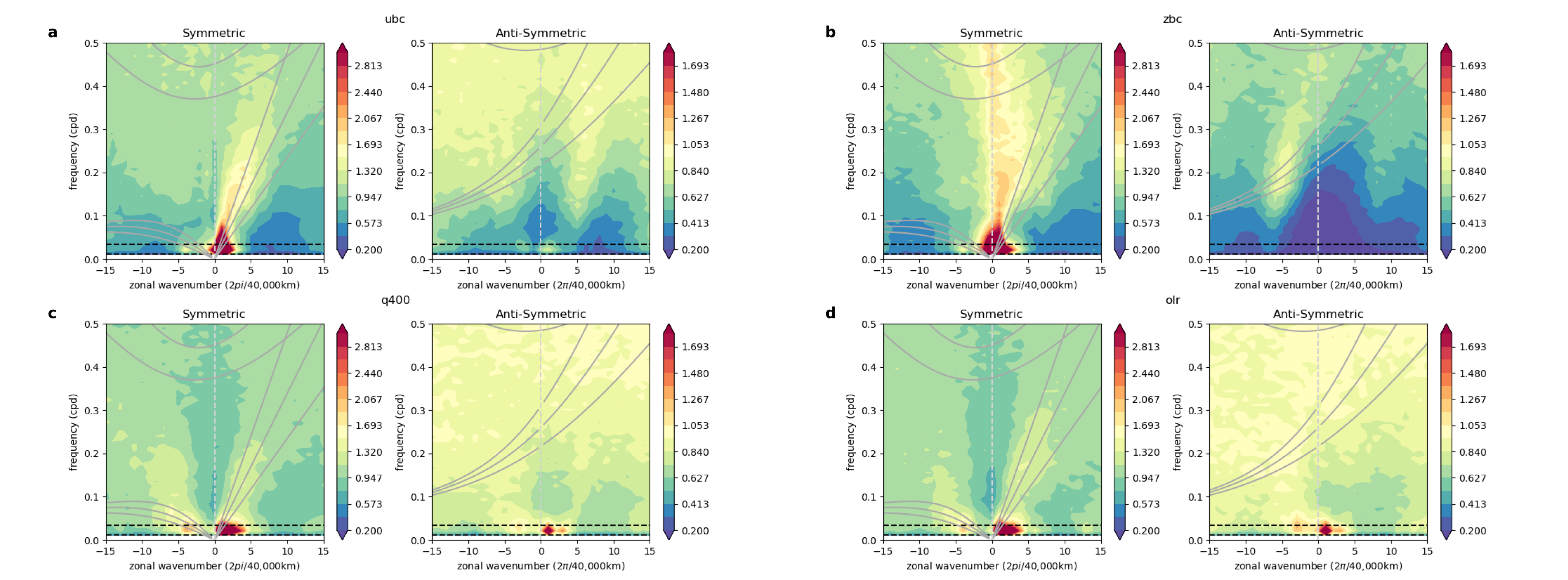}
\caption{Model generated power spectra. 
\textbf{a}, Power spectra of symmetric and asymmetric components 15N-15S, for UBC=U850-U200 (zonal wind velocity). MJO is within band k=1-3, w=0.01-0.03 (30-90 days). Gray lines indicate analytical dispersion curves for major equatorial waves. 
\textbf{b}, Repeated for ZBC=Z850-Z200 (geopotential).
\textbf{c}, Repeated for Q400 (moisture).
\textbf{d}, Repeated for OLR (convection).
Here the model is prompted with observed conditionings from ERA5 1960-2022, as in Fig.~\ref{fig:fig_hovs_model_training}. 
}\label{fig:fig_kw_model_training}
\end{figure*}

Fig.~\ref{fig:fig_kw_model_training} shows the wavenumber-frequency power spectra of the sampled video\cite{wheelerkiladis1999}. The model is able to reproduce the prominent MJO signal (k=1-3, w=0.01-0.03 i.e. 30-90 days) as well as other prominent equatorial waves: Convectively Coupled (CC)-Kelvin and CC-Rossby on symmetric, Mixed Rossby-Gravity (MRG) on asymmetric, inertio-gravity on both\cite{wheelerkiladis1999}. Nevertheless, the model is biased compared with the ERA5 record as it shows for example less pronounced equatorial CC-Kelvin and MRG waves (see SI). The main culprit for this is likely that the low-dimensional conditioning cannot map fully to the high-dimensional variability of observations. Another reason is possibly the presence of non-Gaussian features in the training data leading to suboptimal training\cite{stock2024diffobs} (see SI). More intercomparison is provided in the SI, where we further verify that the MJO is adequately modulated by seasons (doyc, doys) and the ENSO (n34sst). 

In summary the model can reasonably regenerate the ERA5 record, including the MJO and embedded equatorial waves, with nevertheless some biases in representation attributed to low-dimensionality of conditionings and non-Gaussianity.  

\subsection*{Ensemble Sampling}

\begin{figure*}[h!]
\doublespacing 
\centering
\includegraphics[width=1\linewidth]{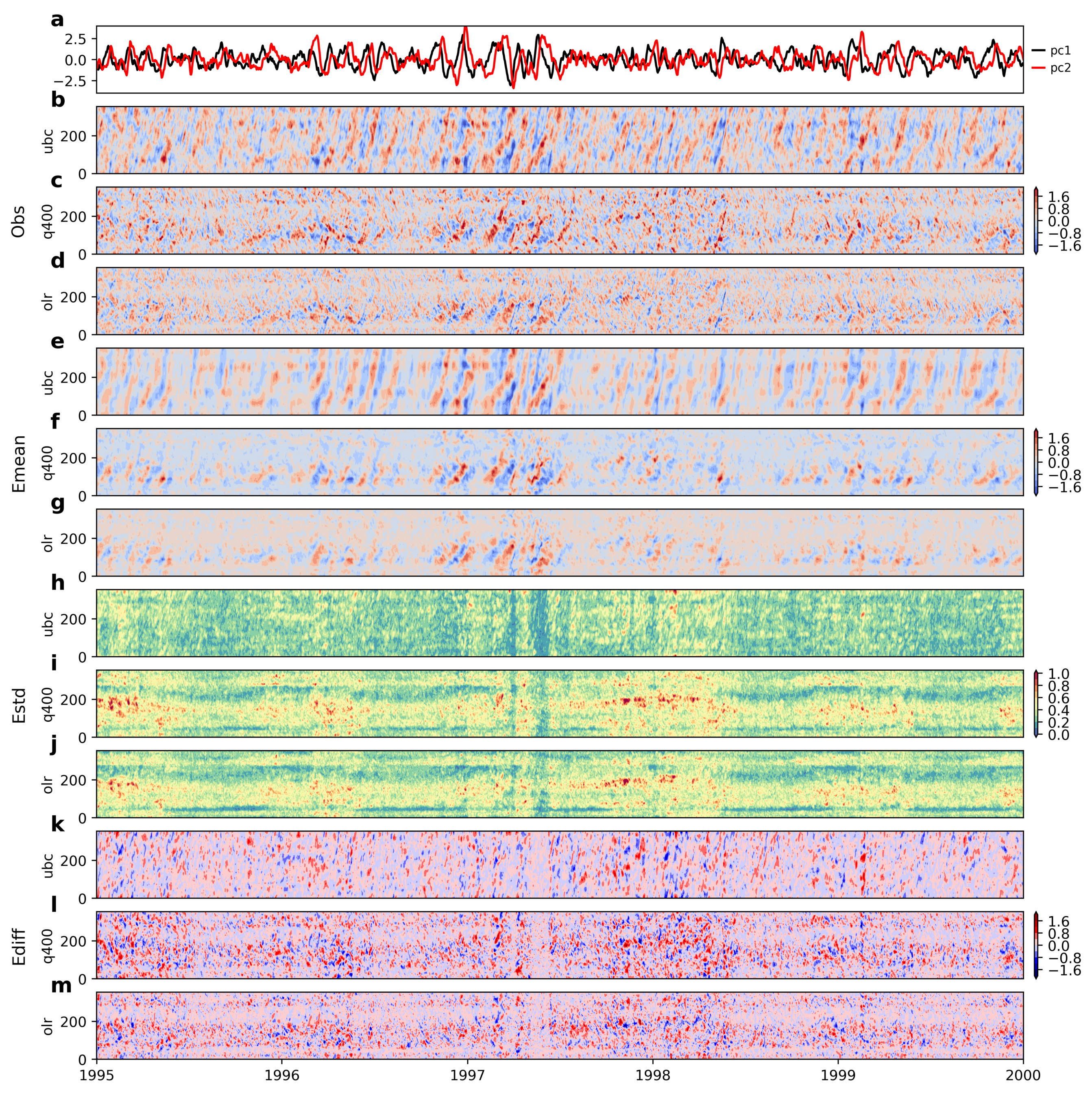}
\caption{Ensemble Sampling. 
\textbf{a}, Conditionings pc1, pc2 as a function of time. 
\textbf{b,c,d}, Hovmollers (15N-15S) for Observed Fields UBC, Q400, OLR from the ERA5 training record, as a function of time and longitude. 
\textbf{e,f,g}, Hovmollers for ensemble mean. 
\textbf{h,i,j}, Hovmollers for ensemble standard deviation.
\textbf{k,l,m}, Hovmollers for ensemble difference, that is difference between one member and the ensemble mean. 
Here the model is prompted with observed ERA5 conditionings 1960-2022, but sampled 10 times. 
}\label{fig:fig_modtrain_ensemblehov}
\end{figure*}

 Here we assess the model's diversity (i.e. randomness) using ensemble sampling. We sample 10 long videos where, for each video, the prompt conditioning is exactly as in the ERA5 observations 1960-2022. 
 This is illustrated in Fig.~{\ref{fig:fig_modtrain_ensemblehov}}. Due to the stochastic nature of the denoising process during sampling, each long video is slightly different\cite{ho2020ddpm, ho2022videodiffusionmodels}. The ensemble mean depicts the smoothed out MJO signal. Its power spectra shows a pronounced MJO signal (and interestingly some CC-Rossby signal), but not other equatorial waves (see SI). 
 The ensemble standard deviation depicts the sample diversity: it shows marked seasonal variations with a maximum in boreal winter in the western to central Pacific warm pool region, consistent with the disorganized convective processes over that region\cite{zhang2005madden}. It also shows variations from one MJO event to the other, and some periods of collapse (no diversity). Finally, we also show ensemble difference, that is the difference between one long video sample and the ensemble mean. The ensemble difference can be marked during some MJO events, leading to slightly modified MJO structures. It also shows eastward and westward propagations that are in fact CC-Kelvin and CC-Rossby waves (as further revealed by power spectra, see SI). These waves are freely generated features, that differ from one sample to the next independent of the prompt (somewhat akin to stochastic noise in low-dimensional MJO models\cite{thual2014stochastic}, but with here organized characteristics).
 
In summary from ensemble analysis, the model generates the MJO with accurate fidelity to prompt, but it also randomly modulates the MJO characteristics and freely generates other equatorial waves.

\section*{Model Prompting}\label{sec_prompting}

\subsection*{Isolated MJO}\label{subsec_prompting_mjos}

Here we explore generating long videos of the MJO, using the sampling strategy from above but now prompting with intentionally more idealized low-dimensional conditionings. This enables deconstructing the underlying processes and assessing in particular the sensitivity to each conditioning in semi-isolation.

\begin{figure*}[h!]
\doublespacing 
\centering
\includegraphics[width=1\linewidth]{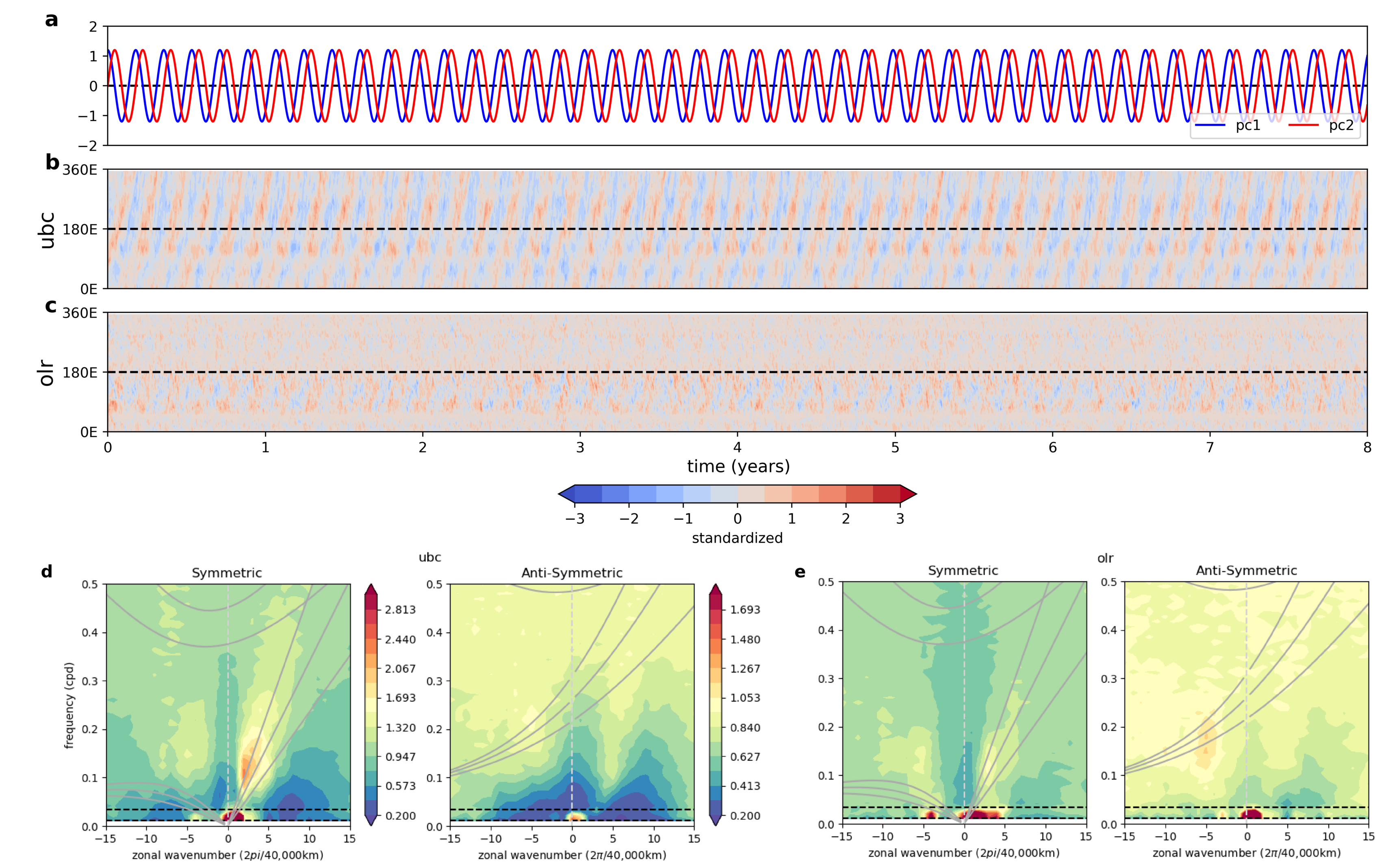}
\caption{Model Prompt for a perpetual MJO. 
\textbf{a}, Conditionings pc1, pc2 as a function of time. 
\textbf{b-c}, Hovmollers (15N-15S) of OLR, UBC as a function of frame (i.e. time) and longitude. 
\textbf{d-e}, Power spectra (symmetric and antisymmetric 15N-15S) for OLR and UBC, following Fig.~\ref{fig:fig_kw_model_training}. 
This uses model version trained for MJO alone (Tab.~\ref{tab:trained_models}). 
}\label{fig:fig_prompt_mjoperio}
\end{figure*}

\begin{figure*}[h!]
\doublespacing 
\centering
\includegraphics[width=1\linewidth]{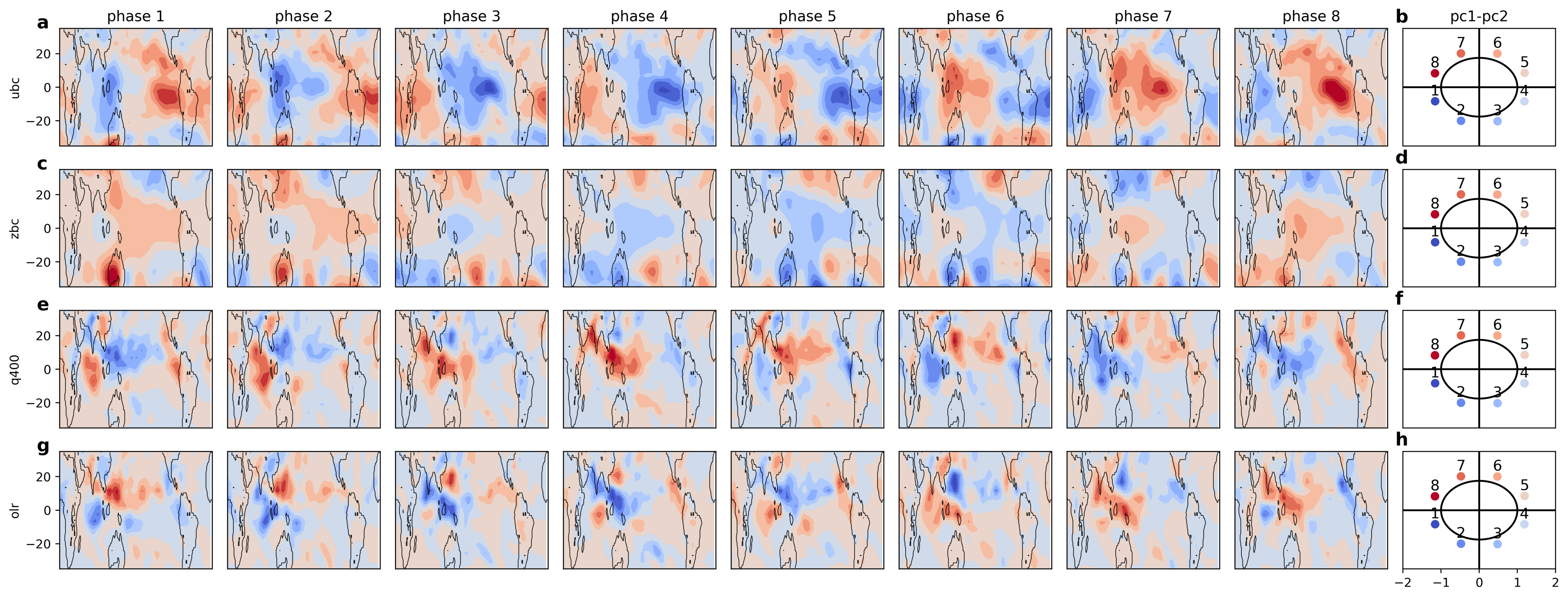}
\caption{MJO Composites.  
\textbf{a, c, e, g}, Composites of UBC, ZBC, Q400 and OLR, by MJO phase 1-8. 
\textbf{b,d,f,h}, Composites of pc1, pc2, by MJO phase 1-8. 
Here the model is prompted for a perpetual MJO as in Fig.~\ref{fig:fig_prompt_mjoperio}. MJO phases are deduced from pc1, pc2.  
This is deduced from the perpetual MJO flow in Fig.~\ref{fig:fig_prompt_mjoperio}
}\label{fig:fig_prompt_mjoperio_compo}
\end{figure*}

The simplest prompt is considering an MJO in isolation i.e. for a model version with no seasonal or ENSO modulation (i.e. as trained on pc1 and pc2 only, see Tab.~\ref{tab:trained_models}). As a simple example we prompt a perpetual MJO, with constant period (65 days) and amplitude (1.2 std), repeating here over a 100 years video for statistical robustness. This is illustrated in Fig.~\ref{fig:fig_prompt_mjoperio}. Despite exhibiting a constant oscillation, the flow here exhibits MJOs with moderate variations in characteristics (e.g. amplitude, phase, structure), and it also exhibits freely generated equatorial waves (CC-Kelvin, CC-Rossby). These stem from the model diversity discussed above. Fig.~\ref{fig:fig_prompt_mjoperio_compo} further shows MJO composites associated with the present flow, as deduced from the RMM-UBC index in the spirt of \cite{wheeler2004rmm}. The composite MJO consistently exhibits eastward propagation and quadrupole structure, consistent with theory\cite{majda2009skeleton} and also retrieved in the ERA5 training data (see SI). This stresses the flow's level of realism despite its idealized conditionings. Another advantage of this flow is its regularity and tractability. 
It may be compared for example to solutions from theoretical MJO models with fixed oscillation period\cite{majda2009skeleton, stechmann2015skeletondata}. It may also be extended to explore parameter sensitivity (e.g. MJO amplitude and period in the conditions) or to generate less trivial MJO sequences (e.g. stalling MJOs, MJO wavetrains, etc\cite{thual2014stochastic, stachnik2015evaluating}), or even intentionally unphysical flows\cite{kamphuis2011retrograde} (see SI). 

\subsection*{Seasonal Modulation}\label{subsec_prompting_seaso}

\begin{figure*}[h!]
\doublespacing 
\centering
\includegraphics[width=1\linewidth]{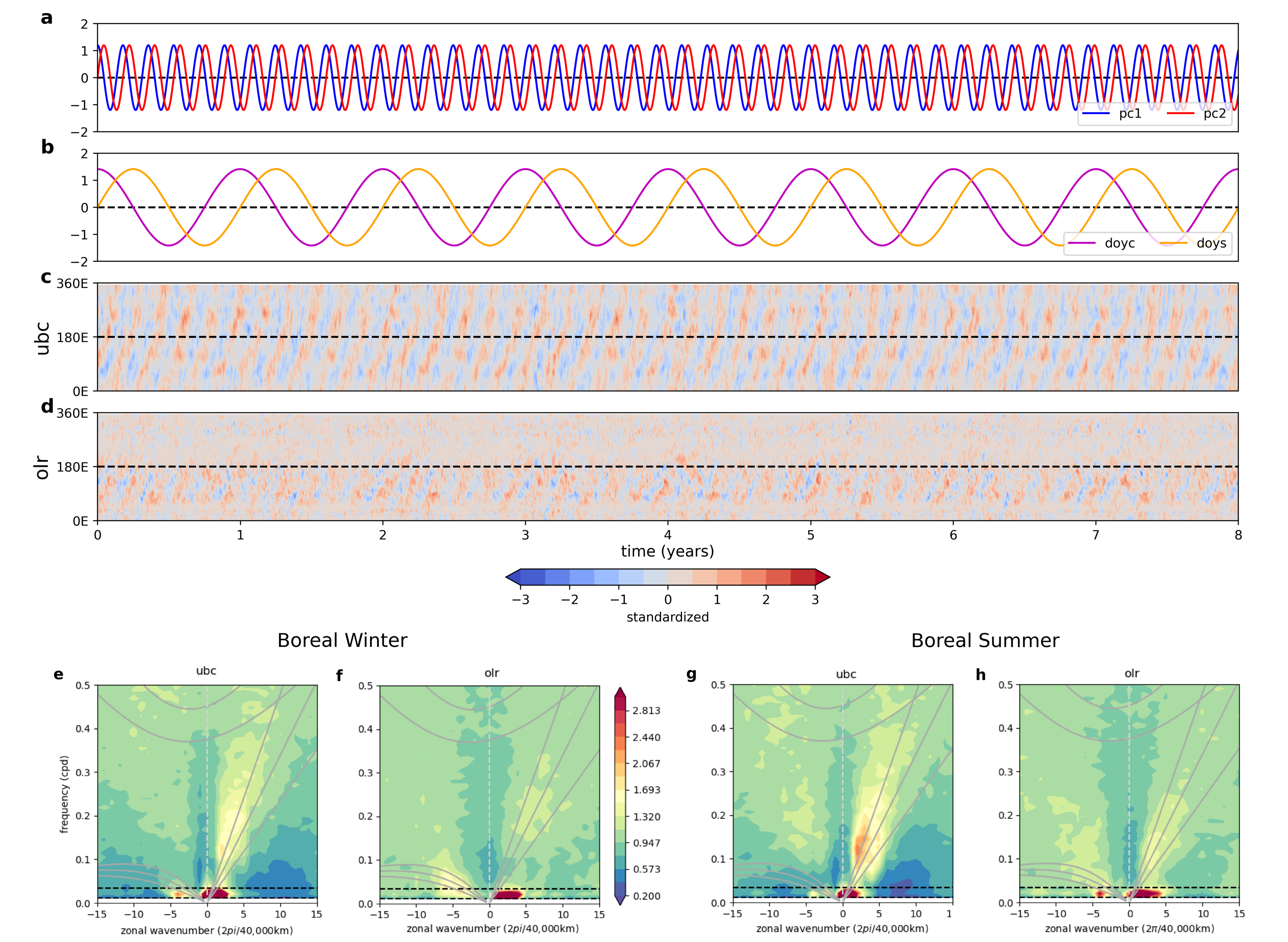}
\caption{Model Prompt for a perpetual MJO with seasonal modulation. 
\textbf{a}, Conditionings pc1, pc2 as a function of time (sinusoides with period 73 days). 
\textbf{b}, Conditionings doyc, doys for seasons (sinusoides with period 1 year).  
\textbf{c-d}, Hovmollers (15N-15S) of OLR, UBC as a function of time and longitude. 
\textbf{e-f}, Seasonal Power spectra (symmetric 15N-15S) for OLR and UBC, with seasonal multiplier centred on boreal winter.
\textbf{g-h}, Seasonal power spectra with seasonal multiplier centred on boreal summer.
This uses model version trained for MJO with Seasons (Tab.~\ref{tab:trained_models}). 
}\label{fig:fig_prompt_mjoseaso}
\end{figure*}

We now consider a prompt with the perpetual MJO from above, but for a model version with seasonal modulation i.e. trained on additional conditionings doyc, doys (see Tab.~\ref{tab:trained_models}). This is illustrated in Fig.~\ref{fig:fig_prompt_mjoseaso} (where for brevity we only show UBC and OLR). By gradually increasing the model conditionings, we are able to assess their role in semi-isolation. 
Here we retrieve the marked seasonal modulation of MJO characteristics, with boreal winter MJOs markedly different in characteristics from the boreal summer MJOs\cite{kikuchi2012bsiso}. In order to distinguish these, we introduce seasonal power spectra that consists in computing regular power spectra on data scaled by a seasonal multiplier (in the spirit of earlier work\cite{masunaga2007kwseaso}, see SI). Here the boreal summer power spectra exhibit a more asymmetric MJO as well as more pronounced CC-Kelvin waves (a features also consistently found in the ERA5 dataset, see SI). This verifies that our low-dimensional embedding (doyc,doys) correctly infers the seasonality in the model. Note that pc1, pc2 in the prompt do not dictate the MJO characteristics alone: in fact, they do not vary between boreal winter and summer, yet the generated characteristics are different. One should be cautions about this interplay between conditionings. For instance, we could alternatively modify the pc1 and pc2 amplitudes with seasons to mimic a slight modulation found in the ERA5 training dataset \cite{wheeler2004rmm} (see SI). Finally, we also considered experiments with perpetual winter or summer (i.e. constant doyc, doys values), leading to similar results (see SI).

\subsection*{ENSO Modulation}\label{subsec_prompting_enso}

\begin{figure*}[h!]
\doublespacing 
\centering
\includegraphics[width=1\linewidth]{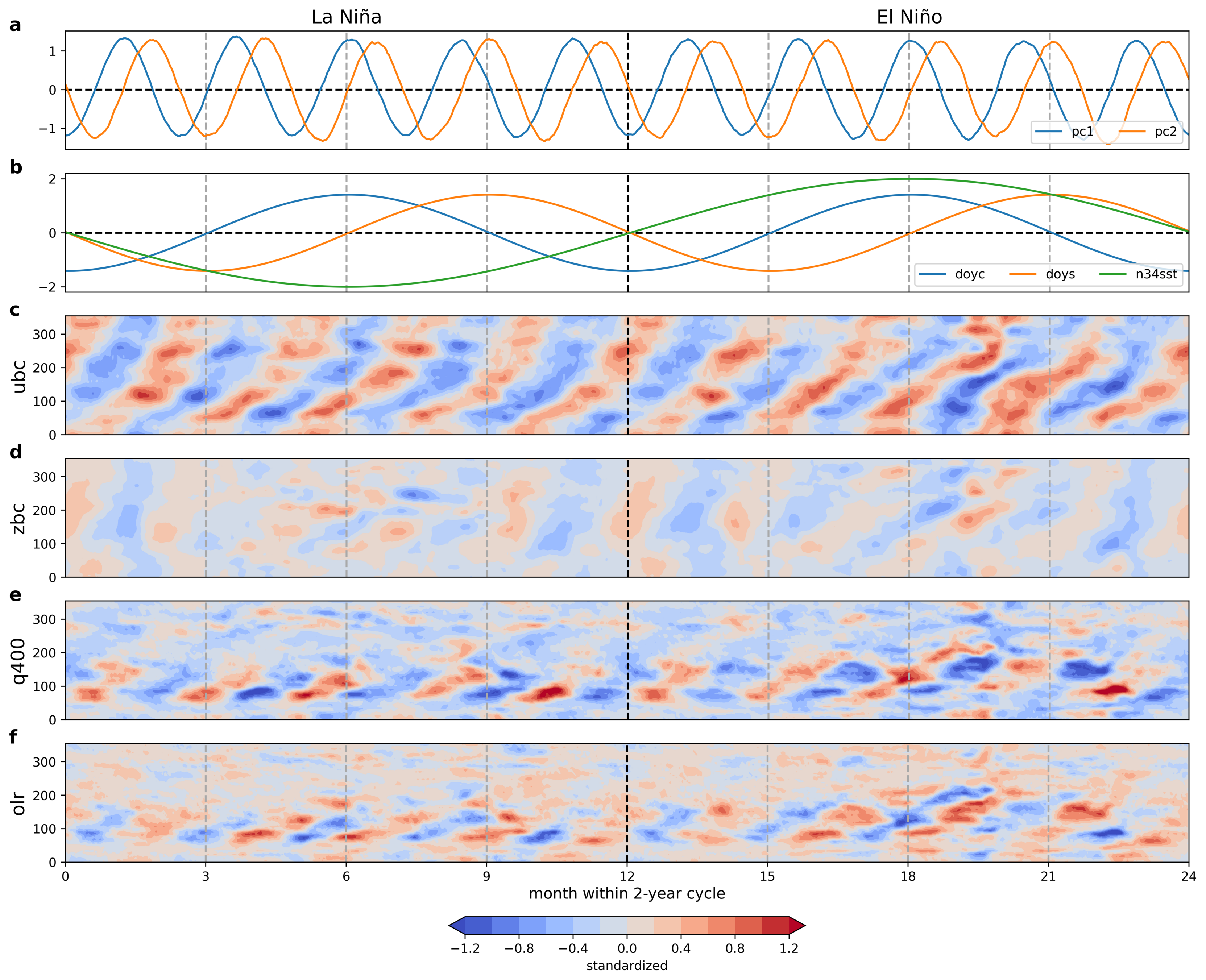}
\caption{Model Prompt for a perpetual MJO with seasonal and ENSO modulation.   
\textbf{a}, Conditionings pc1, pc2 as a function of time (sinusoides with period 73 days). 
\textbf{b}, Conditionings doyc, doys for seasons (sinusoides with period 1 year), and n34sst for ENSO (sinusoide with period 2 years). Note all conditionings are 2-year periodic. 
\textbf{c-f}, Hovmollers (15N-15S) of Phase-Composite for UBC, ZBC, Q400, OLR, as a function of 2-year phase and longitude. Phase Composites are binned averages along the 2-year phase, which extracts the periodic signals. 
This uses model version trained for MJO with Seasons/ENSO (Tab.~\ref{tab:trained_models}). 
}\label{fig:fig_prompt_mjoenso}
\end{figure*}

We now further increase the model prompt complexity by adding ENSO modulation, i.e. the conditioning n34sst (see Tab.~\ref{tab:trained_models}). This is illustrated in Fig.~\ref{fig:fig_prompt_mjoenso}. As a simple example, the ENSO cycle is here a sinusoide with period 2 years where both El Ni\~no and La Ni\~na peak in boreal winter (doyc=1, doys=0), and where the perpetual MJO (pc1, pc2) has a 73 days period (see SI). The present prompt is highly oversimplified compared to nature where both the ENSO and MJO are highly irregular\cite{capotondi2020book, zhang2005madden}, but greatly simplifies the analysis. In fact, all conditionings are here 2-year periodic and there are exactly 5 MJOs with identical timing each El Ni\~no or La Ni\~na years. Thus we can easily compare the structure of the MJOs during El Ni\~no and La Ni\~na, using for example phase composites (respective to the 2 year phase). The composites in Fig.~\ref{fig:fig_prompt_mjoenso} show increased intensity, span and eastward extent of MJOs in the aftermath of El Ni\~nos (months 18 to 21 in Fig.~\ref{fig:fig_prompt_mjoenso}) compared to La Ni\~nas (months 6 to 9). This is a well observed feature\cite{hendon2007ensomjo} that the video diffusion model captures in its more idealized setup. Once again, we find that the RMM index (pc1, pc2) dictates the MJO characteristics (structure, intensity and timing) in tandem with the other conditionings (seasonality, ENSO state), rather than alone. This is evident in Fig.~\ref{fig:fig_prompt_mjoenso} where the MJOs differ during El Ni\~no and La Ni\~na  despite identical pc1, pc2. 

In summary, the present video diffusion model can generate intentionally idealized MJOs (e.g. a perpetual periodic MJO), which decouples processes and allows for more tractable analysis. By gradually incorporating conditionings, we can assess their role in semi-isolation. We find in fact that while the RMM index (pc1, pc2) mostly dictates the MJO characteristics (structure, intensity and timing), it does so in tandem with the other conditionings.

\section*{Discussion}\label{sec_conclusions}

In the present paper, we have trained and sampled a video diffusion model for the MJO. The model is trained on the ERA5 atmospheric reanalysis, then sampled to generate long videos of MJO sequences from low-dimensional conditioning. The model can generate a reasonably realistic MJOs despite some biases in representation. Prompting the model with intentionally idealized conditionings decouples processes and allows for more tractable analysis, e.g. assessing the role of seasonal or ENSO modulations in isolation. Importantly, the model's semi-realism extends to these more idealized prompts, which provides physical insight. The approach also enables quick iterations: in our setup it takes around 30 minutes to generate a 60 years MJO record, which is roughly on par with intermediate complexity models in terms of computing time. 

The present approach provides a tractable link between generative modeling\cite{price2025gencastera5, stock2024diffobs, martin2022xai, shin2024deep} and theoretical understanding\cite{majda2009skeleton,stechmann2015skeletondata,thual2014stochastic, stachnik2015evaluating}, one advantage being that key climate metrics may be mapped more directly to a realistic high-dimensional flow. For instance, the present method may outperform traditional statistical methods (e.g. MJO composites, principal components\cite{wheeler2004rmm}) at reconstructing details embedded within the MJO (e.g. CC-Kelin or CC-Rossby waves\cite{wheelerkiladis1999}). As another instance, restricting the model conditionings to low-dimensional tokens may simplify the generative process in terms of experimentation and interpretation\cite{stechmann2015skeletondata}. Nevertheless it remains to be determined if this approach has some merits in predictive settings\cite{delaunaychristensen2022mjodl}. More generally, the present results support the notion that deep learning models may bring physical insight, and are thus not just mere black boxes\cite{martin2022xai, shin2024deep}. 

As a perspective to the present work, we may improve the present video diffusion model setup. The model in fact shows biases in representing the MJO that hinder interpretation to some extent. For instance, the ERA5 training dataset exhibits non-Gaussian features that may degrade training performances (see SI), but could be mitigated by suitable data rescaling (e.g. power transforms, quantile transforms, etc). Sensitivity to model parameters could be assessed more systematically, although we observed no major sensitivity from brief testing. 
As another perspective, one may quantify which low-dimensional (i.e. latent) variables are most suitable to prompt the MJO variability and its modulation. Some key ingredients may in fact be better be captured by other metrics\cite{stechmann2015skeletondata, stachnik2015evaluating, capotondi2020book}. 
More generally, the present approach may also be extended to other climate problems where low-dimensional theory is established, but requires linking to spatio-temporal complexity\cite{capotondi2020book, ren2025videodiffcyclones}. Ultimately, the present "climate prompting" approach may clarify the relationship between planetary-scale MJO dynamics, embedded convectively coupled equatorial waves and external modulations.


\pdfbookmark[1]{Methods}{section:sec_methods}
\section*{Methods}\label{sec_methods}

\subsection*{Training Dataset}

The training dataset for our model is from the ECMWF Reanalysis v5 (ERA5), a global atmospheric reanalysis\cite{hersbach2020era5}, as retrieved here from WeatherBench\cite{rasp2024weatherbench2}.
The domain is $35^{\circ}N-35^{\circ}S$ at resolution 5.625x4.375 (64x16 grid), which covers the tropical area with planetary scale resolution. Data is daily covering here 1960-2022, with 365 days per years (as we remove Feb 29th on leap years). We select a few atmospheric variables from ERA5: zonal wind velocity at 850 and 200 hPa (U850, U200, in $m.s^{-1}$), geopotential at 850 and 200 hPa (Z850, Z200 in $m^{2}.s^{-2}$), specific humidity at 400 hPa (Q400 in $kg.kg^{-1}$), and outgoing long wave radiation (OLR, as deduced from mean top net long wave radiation flux in ERA5, in $W.m^{-2}$). We also deduce the fields UBC=U850-U200 and ZBC=Z850-Z200 for first baroclinic mode motion in the atmosphere.  With this, we consider a more compact dataset, representative of the MJO, that consists of the fields UBC, ZBC, Q400, OLR\cite{stechmann2015skeletondata}. 
From the raw regridded fields, we remove the daily climatology, then remove interannual variability using a Butterworth high-pass filter with cutoff 120 days, then remove residual daily climatology again (see SI). This extracts intraseasonal anomalies in the spirit of the original index\cite{wheeler2004rmm}, but more consistently ensures a zero-mean and temporal smoothness (as our goal is MJO generation but not prediction). 
%

Conditioning metrics are low-dimensional and time-dependent. They are listed in Tab.~\ref{tab:conditionings}.  
A first pair of conditionings is a slightly modified RMM index representative of the MJO\cite{wheeler2004rmm, stachnik2015evaluating}, denoted hereafter as RMM-UBC. The RMM-UBC index consists of the two first principal components deduced from the 15N-15S concatenated fields UBC and OLR. Its advantage here is to be directly deducible from the model output fields, while it remains very similar to the original RMM index (see SI). To deduce the principal components from a given model output fields, we project the fields onto the original spatial structures (or empirical orthogonal functions). This leads to two conditionings, denoted hereafter as pc1, pc2, that embed the MJO phase and amplitude. 
A second pair of conditionings is a sinusoidal embedding of seasons, defined as $\text{doyc}=\cos(2\pi \text{doy}/365)$ and $\text{doys}=\sin(2\pi \text{doy}/365)$, where doy is the day of the year (spanning exactly 365 days in our dataset). This embeds the seasons in minimal fashion, given that they modulate the MJO and tropical intraseasonal variability in general\cite{kikuchi2012bsiso}. 
A third conditioning is the index Nino 3.4 SST, denoted hereafter as n34sst, obtained from ERSSTv5 dataset and interpolated from monthly to daily sampling\cite{huang2017ersstv5}, that embeds the state of the ENSO. In fact the ENSO also modulates MJO activity\cite{hendon2007ensomjo}. 

In summary the ERA5 training dataset includes sample fields UBC, ZBC, Q400 and OLR (that depend on longitude, latitude and time), and conditionings are pc1, pc2, doyc, doys and n34sst (that depend on time). 

\subsection*{Video Diffusion Model}

Diffusion models learn to map a simple prior distribution, typically Gaussian noise, to a complex data distribution by reversing a stochastic forward process through iterative denoising steps\cite{ho2020ddpm}. While widely popularized for image generation\cite{saharia2022imagen}, diffusion models have recently been adapted to simulate complex physical systems and climate dynamics\cite{price2025gencastera5, stock2024diffobs, ren2025videodiffcyclones}.

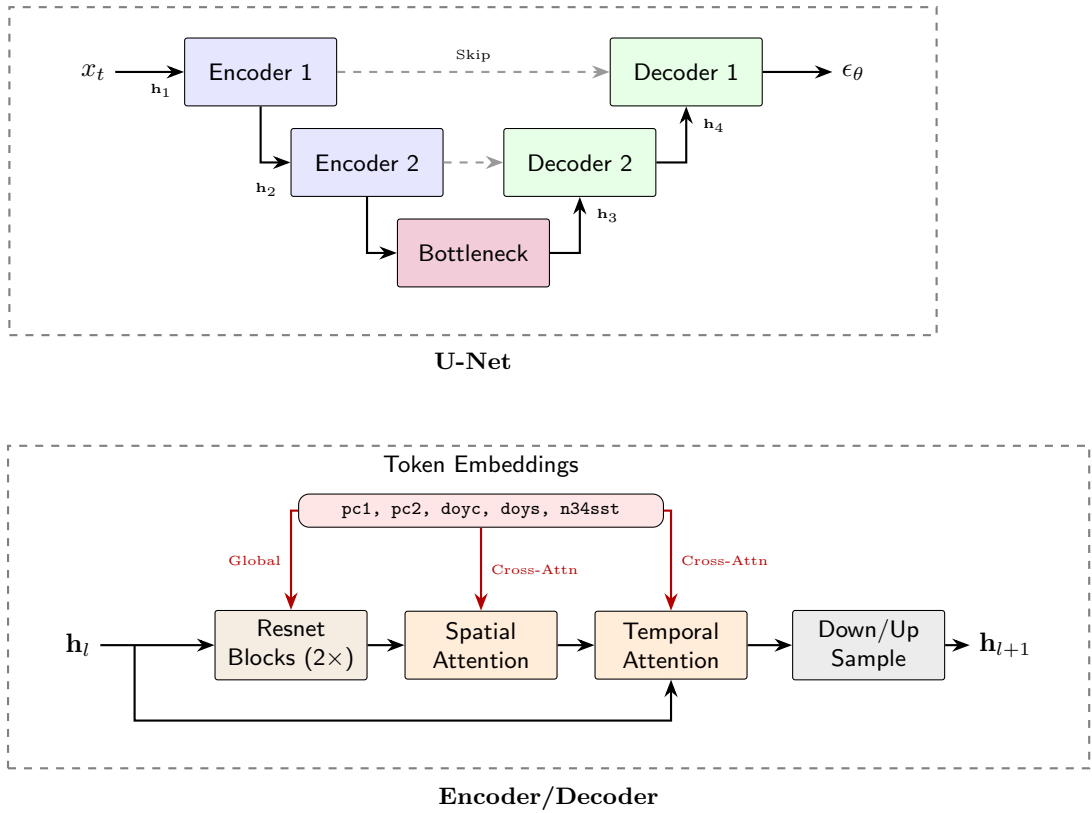
\begin{figure}[ht]
\centering
\begin{tikzpicture}[
node distance=0.5cm and 1.0cm,
>=Stealth,
unode/.style={draw, rectangle, rounded corners=1pt, minimum width=2.0cm, minimum height=0.9cm, font=\small\sffamily, align=center},
enc/.style={unode, fill=blue!10},
dec/.style={unode, fill=green!10},
btl/.style={unode, fill=purple!20},
box/.style={unode}, 
resnet/.style={box, fill=brown!15},
attn/.style={box, fill=orange!15},
sampling/.style={box, fill=gray!15},
prompt/.style={draw, rounded corners, fill=red!10, font=\footnotesize\ttfamily, minimum width=4.8cm}
]
\begin{scope}[local bounding box=unet_scope]
\node (xt) at (-5.0,0) {$x_t$};
\node[enc] (ge1) at (-2.8,0) {Encoder 1};
\node[enc] (ge2) at (-1.4,-1.2) {Encoder 2};
\node[btl] (gb) at (0,-2.4) {Bottleneck};
\node[dec] (gd2) at (1.4,-1.2) {Decoder 2};
\node[dec] (gd1) at (2.8,0) {Decoder 1};
\node (eps) at (5.0,0) {$\epsilon_\theta$};
\draw[->, thick] (xt) -- (ge1);
\draw[->, thick] (ge1) |- (ge2);
\draw[->, thick] (ge2) |- (gb);
\draw[->, thick] (gb) -| (gd2);
\draw[->, thick] (gd2) -| (gd1);
\draw[->, thick] (gd1) -- (eps);
\draw[->, dashed, gray!80, thick] (ge1) -- (gd1) node[midway, above, font=\tiny, text=black] {Skip};
\draw[->, dashed, gray!80, thick] (ge2) -- (gd2);
\node[font=\tiny, below left=0.05cm of ge1.west] {$\mathbf{h}_1$};
\node[font=\tiny, below left=0.15cm and 0.05cm of ge2.west] {$\mathbf{h}_2$};
\node[font=\tiny, right=0.1cm of gd2.south, yshift=-0.25cm] {$\mathbf{h}_3$};
\node[font=\tiny, right=0.1cm of gd1.south, yshift=-0.25cm] {$\mathbf{h}_4$};
\end{scope}
\node[draw, gray, thick, dashed, inner sep=18pt, fit={(xt) (eps) (gb)}, minimum width=12.2cm] (frame1) {};
\node[below=0.1cm of frame1, font=\small\bfseries] {U-Net};
\begin{scope}[yshift=-7.6cm, local bounding box=block_scope]
\node (in) at (-5.2,0) {$\mathbf{h}_l$};
\node[resnet] (res) at (-2.4,0) {Resnet \\ Blocks (2$\times$)};
\node[attn] (spatial) at (0.1,0) {Spatial \\ Attention};
\node[attn] (temporal) at (2.6,0) {Temporal \\ Attention};
\node[sampling] (sample) at (5.2,0) {Down/Up \\ Sample};
\node (out) at (7.0,0) {$\mathbf{h}_{l+1}$};
\draw[->, thick] (in) -- (res);
\draw[->, thick] (res) -- (spatial);
\draw[->, thick] (spatial) -- (temporal);
\draw[->, thick] (temporal) -- (sample);
\draw[->, thick] (sample) -- (out);
\node[prompt, above=1.1cm of spatial] (p_box) {pc1, pc2, doyc, doys, n34sst};
\node[above=0.1cm of p_box, font=\small\sffamily] {Token Embeddings};
\draw[->, thick, red!70!black] (p_box.west) -| (res.north) node[pos=0.75, left, font=\tiny] {Global};
\draw[->, thick, red!70!black] (p_box.south) -- (spatial.north) node[midway, right, font=\tiny] {Cross-Attn};
\draw[->, thick, red!70!black] (p_box.east) -| (temporal.north) node[pos=0.75, right, font=\tiny] {Cross-Attn};
\draw[->, thick] ($(in.east)!0.3!(res.west)$) -- ++(0,-1.0) coordinate (arrow_bottom) -| (temporal.south);
\end{scope}
\node[draw, gray, thick, dashed, inner sep=18pt, fit={(in) (out) (p_box) (arrow_bottom)}, minimum width=12.2cm] (frame2) {};
\node[below=0.1cm of frame2, font=\small\bfseries] {Encoder/Decoder};
\end{tikzpicture}
\caption{\textbf{Architecture of the video diffusion model.} The model predicts the noise $\epsilon_\theta$ added to the input atmospheric fields $x_t$. It implements a 3D hierarchical U-Net architecture with tokens (conditionings) embedded within the Encoder/Decoder blocks via global and cross-attention pathways. Skip connections concatenate encoder output with corresponding decoder input. Encoders downsample while Decoders upsample the feature maps.}
\label{fig:unet}
\end{figure}

The present video model's architecture is a symmetric U-Net integrated with transformers\cite{vaswani2023attentionneed}, following the configuration of Imagen\cite{saharia2022imagen} but extended with a temporal dimension\cite{ho2022videodiffusionmodels}. Its architecture is illustrated in Fig.~\ref{fig:unet}. We decouple the attention mechanisms within the Encoders and Decoders into distinct spatial and temporal blocks: spatial attention processes frames independently, while temporal attention batches spatial coordinates to compute self-attention across the sequence\cite{bastek2023metamaterials}. A residual skip connection bridges the block input directly to the temporal attention layer to maintain feature stability. Low-dimensional conditionings are integrated via two distinct pathways: a global pathway where climate indices are embedded within the Resnet blocks, and a cross-attention pathway where they act as sequence tokens for both spatial and temporal attention. This dual structure enforces both physical consistency and temporal coherence.

We consider various trained models, as listed in Tab.\ref{tab:trained_models}, that differ by conditionings used. At minimum the model embeds only MJO conditionings, and at most it integrates all conditionings. Iterating on these allows us to isolate the role of each conditioning. We also tested model versions with less fields (e.g. UBC and OLR) for quick prototyping (not shown).
Training samples are sequences of 16 frames (i.e. days) randomly selected from the ERA5 dataset, thus with dimensions [lon, lat, frame] for fields, and [frame] for conditionings. Considering 16 frames here achieves a reasonable balance between temporal coherence and computation cost. The sample fields and conditionings are normalized for model training, and for clarity they are also systematically standardized in the paper's figures. When normalizing or standardizing we systematically use rescaling parameters (mean, standard deviation, min, max) deduced from the full ERA5 training dataset. 

After training, we may sample the model: we input arbitrary conditionings (each 16 frames) which generates a  short sequence (with identical format as the training samples). To generate longer sequences spawning multiple years, we combine multiple samples generated by the model using Brick-Wall Denoising\cite{yuan2025brickwall}. The approach iteratively shifts denoising windows during the sampling process, which effectively mixes overlapping samples into a longer sequence. While it may introduce minor artifacts, it effectively maintains temporal coherence across the full sequence. The method is here preferred for its relative simplicity, although many other approaches exists\cite{xie2025regressivevideo,cachay2025rollingdiff}.

Model parameters are listed in Tab.~\ref{tab:model_hyperparameters} 
Models are trained for 20000 steps, with condition dropout 0.1 and dynamic thresholding at 99th quantile.
Sampling uses Denoising Diffusion Implicit Models\cite{song2020ddim} using 250 timesteps, with full stochasticity ($\eta=1$ in their paper) and no guidance. 
Brick-Wall Denoising uses a stride of 3 frames\cite{yuan2025brickwall}. 
The models were trained on Palmetto Cluster at Clemson University\cite{antao2024palmetto} (2 GPUs A100, 64 CPU cores, 250Gb memory). One model training takes roughly 12 hours, and sampling a 60 years video takes around 30 minutes.


\backmatter
\begin{appendices}


\hypersetup{bookmarksdepth=-1}

\bmhead{Data availability}
\begin{itemize}
    \item The present ERA5 dataset version was sourced from WeatherBench\cite{rasp2024weatherbench2} (\url{https://github.com/google-research/weatherbench2}).
    \item The Niño 3.4 index was sourced from the NOAA Physical Sciences Laboratory (\url{https://psl.noaa.gov/data/timeseries/month/DS/Nino34_CPC/}).
    \item The video diffusion model code (pytorch) is adapted from Bastek et al. 2023\cite{bastek2023metamaterials}.
    \item The wavenumber-frequency power spectra are computed using code from \url{https://github.com/brianpm/wavenumber_frequency}. 
\end{itemize}


\bmhead{Authors' contributions} 
S.T. and F.L. designed research. S.T. performed research. All authors discussed research and wrote the paper. 

\bmhead{Competing interests}
The authors declare no conflict of interest.

\hypersetup{bookmarksdepth} 


\clearpage
\pdfbookmark[1]{References}{bibliography}
\bibliography{BIB}

@misc{antao2024palmetto,
  title={Modernizing Clemson University's Palmetto Cluster: Lessons Learned from 17 Years of HPC Administration},
  author={Antao, Asher and Burton, James Daly and Dawson, Douglas and Gemmill, Jill and Gerstener, Zachary and Godfrey, Ben and Groel, Scott and Jordan, Zach and Ligon, Becky and Smith, Dane and others},
  booktitle={Practice and Experience in Advanced Research Computing 2024: Human Powered Computing},
  pages={1--9},
  year={2024}
}

@misc{vaswani2023attentionneed,
      title={Attention Is All You Need}, 
      author={Ashish Vaswani and Noam Shazeer and Niki Parmar and Jakob Uszkoreit and Llion Jones and Aidan N. Gomez and Lukasz Kaiser and Illia Polosukhin},
      year={2023},
      eprint={1706.03762},
      archivePrefix={arXiv},
      primaryClass={cs.CL},
      url={https://arxiv.org/abs/1706.03762}, 
}

@msic{bastek2023metamaterials,
  title={Inverse design of nonlinear mechanical metamaterials via video denoising diffusion models},
  author={Bastek, Jan-Hendrik and Kochmann, Dennis M},
  journal={Nature Machine Intelligence},
  volume={5},
  number={12},
  pages={1466--1475},
  year={2023},
  publisher={Nature Publishing Group UK London}
}

@misc{ho2020ddpm,
      title={Denoising Diffusion Probabilistic Models}, 
      author={Jonathan Ho and Ajay Jain and Pieter Abbeel},
      year={2020},
      eprint={2006.11239},
      archivePrefix={arXiv},
      primaryClass={cs.LG},
      url={https://arxiv.org/abs/2006.11239}, 
}

@misc{ho2022videodiffusionmodels,
      title={Video Diffusion Models}, 
      author={Jonathan Ho and Tim Salimans and Alexey Gritsenko and William Chan and Mohammad Norouzi and David J. Fleet},
      year={2022},
      eprint={2204.03458},
      archivePrefix={arXiv},
      primaryClass={cs.CV},
      url={https://arxiv.org/abs/2204.03458}, 
}

@misc{saharia2022imagen,
      title={Photorealistic Text-to-Image Diffusion Models with Deep Language Understanding}, 
      author={Chitwan Saharia and William Chan and Saurabh Saxena and Lala Li and Jay Whang and Emily Denton and Seyed Kamyar Seyed Ghasemipour and Burcu Karagol Ayan and S. Sara Mahdavi and Rapha Gontijo Lopes and Tim Salimans and Jonathan Ho and David J Fleet and Mohammad Norouzi},
      year={2022},
      eprint={2205.11487},
      archivePrefix={arXiv},
      primaryClass={cs.CV},
      url={https://arxiv.org/abs/2205.11487}, 
}

@article{song2020ddim,
  title={Denoising diffusion implicit models},
  author={Song, Jiaming and Meng, Chenlin and Ermon, Stefano},
  journal={arXiv preprint arXiv:2010.02502},
  year={2020}
}

@inproceedings{yuan2025brickwall,
  title={Brick-Diffusion: Generating Long Videos with Brick-to-Wall Denoising},
  author={Yuan, Yunlong and Guo, Yuanfan and Wang, Chunwei and Xu, Hang and Zhang, Li},
  booktitle={ICASSP 2025-2025 IEEE International Conference on Acoustics, Speech and Signal Processing (ICASSP)},
  pages={1--5},
  year={2025},
  organization={IEEE}
}

@inproceedings{xie2025regressivevideo,
  title={Progressive autoregressive video diffusion models},
  author={Xie, Desai and Xu, Zhan and Hong, Yicong and Tan, Hao and Liu, Difan and Liu, Feng and Kaufman, Arie and Zhou, Yang},
  booktitle={Proceedings of the Computer Vision and Pattern Recognition Conference},
  pages={6322--6332},
  year={2025}
}

@article{behrens2022non,
  title={Non-linear dimensionality reduction with a variational encoder decoder to understand convective processes in climate models},
  author={Behrens, Gunnar and Beucler, Tom and Gentine, Pierre and Iglesias-Suarez, Fernando and Pritchard, Michael and Eyring, Veronika},
  journal={Journal of Advances in Modeling Earth Systems},
  volume={14},
  number={8},
  pages={e2022MS003130},
  year={2022},
  publisher={Wiley Online Library}
}

@article{bi2023panguweatherera5,
  title={Accurate medium-range global weather forecasting with 3D neural networks},
  author={Bi, Kaifeng and Xie, Lingxi and Zhang, Hengheng and Chen, Xin and Gu, Xiaotao and Tian, Qi},
  journal={Nature},
  volume={619},
  number={7970},
  pages={533--538},
  year={2023},
  publisher={Nature Publishing Group}
}

@article{cachay2025rollingdiff,
  title={Elucidated Rolling Diffusion Models for Probabilistic Weather Forecasting},
  author={Cachay, Salva R{\"u}hling and Aittala, Miika and Kreis, Karsten and Brenowitz, Noah and Vahdat, Arash and Mardani, Morteza and Yu, Rose},
  journal={arXiv preprint arXiv:2506.20024},
  year={2025}
}

@article{chen2023fuxi,
  title={FuXi: A cascade machine learning forecasting system for 15-day global weather forecast},
  author={Chen, Lei and Zhong, Xiaohui and Zhang, Feng and Cheng, Yuan and Xu, Yinghui and Qi, Yuan and Li, Hao},
  journal={npj climate and atmospheric science},
  volume={6},
  number={1},
  pages={190},
  year={2023},
  publisher={Nature Publishing Group UK London}
}

@article{delaunaychristensen2022mjodl,
  title={Interpretable deep learning for probabilistic MJO prediction},
  author={Delaunay, Antoine and Christensen, Hannah M},
  journal={Geophysical Research Letters},
  volume={49},
  number={16},
  pages={e2022GL098566},
  year={2022},
  publisher={Wiley Online Library}
}

@article{kashinath2021physics,
  title={Physics-informed machine learning: case studies for weather and climate modelling},
  author={Kashinath, Karthik and Mustafa, Mustafa and Albert, Adrian and Wu, Jean-Luc and Jiang, C and Esmaeilzadeh, Soheil and Azizzadenesheli, Kamyar and Wang, R and Chattopadhyay, Ashesh and Singh, Aakanksha and others},
  journal={Philosophical Transactions of the Royal Society A: Mathematical, Physical and Engineering Sciences},
  volume={379},
  number={2194},
  year={2021},
  publisher={The Royal Society}
}

@article{lam2023graphcastera5,
  title={Learning skillful medium-range global weather forecasting},
  author={Lam, Remi and Sanchez-Gonzalez, Alvaro and Willson, Matthew and Wirnsberger, Peter and Fortunato, Meire and Alet, Ferran and Ravuri, Suman and Ewalds, Timo and Eaton-Rosen, Zach and Hu, Weihua and others},
  journal={Science},
  volume={382},
  number={6677},
  pages={1416--1421},
  year={2023},
  publisher={American Association for the Advancement of Science}
}

@article{martin2022xai,
  title={Using simple, explainable neural networks to predict the Madden-Julian oscillation},
  author={Martin, Zane K and Barnes, Elizabeth A and Maloney, Eric},
  journal={Journal of Advances in Modeling Earth Systems},
  volume={14},
  number={5},
  pages={e2021MS002774},
  year={2022},
  publisher={Wiley Online Library}
}

@misc{pathak2022fourcastnet,
      title={FourCastNet: A Global Data-driven High-resolution Weather Model using Adaptive Fourier Neural Operators}, 
      author={Jaideep Pathak and Shashank Subramanian and Peter Harrington and Sanjeev Raja and Ashesh Chattopadhyay and Morteza Mardani and Thorsten Kurth and David Hall and Zongyi Li and Kamyar Azizzadenesheli and Pedram Hassanzadeh and Karthik Kashinath and Animashree Anandkumar},
      year={2022},
      eprint={2202.11214},
      archivePrefix={arXiv},
      primaryClass={physics.ao-ph},
      url={https://arxiv.org/abs/2202.11214}, 
}

@article{price2025gencastera5,
  title={Probabilistic weather forecasting with machine learning},
  author={Price, Ilan and Sanchez-Gonzalez, Alvaro and Alet, Ferran and Andersson, Tom R and El-Kadi, Andrew and Masters, Dominic and Ewalds, Timo and Stott, Jacklynn and Mohamed, Shakir and Battaglia, Peter and others},
  journal={Nature},
  volume={637},
  number={8044},
  pages={84--90},
  year={2025},
  publisher={Nature Publishing Group}
}

@article{ren2025videodiffcyclones,
  title={Improving Tropical Cyclone Forecasting With Video Diffusion Models},
  author={Ren, Zhibo and Nath, Pritthijit and Shukla, Pancham},
  journal={arXiv preprint arXiv:2501.16003},
  year={2025}
}

@article{shin2024deep,
  title={Deep learning reveals moisture as the primary predictability source of MJO},
  author={Shin, Na-Yeon and Kim, Daehyun and Kang, Daehyun and Kim, Hyemi and Kug, Jong-Seong},
  journal={npj Climate and Atmospheric Science},
  volume={7},
  number={1},
  pages={11},
  year={2024},
  publisher={Nature Publishing Group UK London}
}

@article{stock2024diffobs,
  title={Diffobs: Generative diffusion for global forecasting of satellite observations},
  author={Stock, Jason and Pathak, Jaideep and Cohen, Yair and Pritchard, Mike and Garg, Piyush and Durran, Dale and Mardani, Morteza and Brenowitz, Noah},
  journal={arXiv preprint arXiv:2404.06517},
  year={2024}
}

@article{yao2024machine,
  title={Machine Learning Models Use Large Scale Signals to Forecast the MJO},
  author={Yao, Lin and Yang, Da and Duncan, James and Chattopadhyay, Ashesh Kumar and Hassanzadeh, Pedram and Bhimji, Wahid and Yu, Bin},
  journal={European Geosciences Union General Assembly 2024 (EGU24)},
  pages={20993},
  year={2024}
}

@inproceedings{capotondi2020book,
  author = {Capotondi, A. and Wittenberg, A. and Kug, J.-S. and Takahashi, K. and PcPhaden, M. },
  title = {ENSO diversity},
  editor = {McPhaden, M.J. and Santoso, A. and Cai, W.},
  volume = {253},
  booktitle = {El Ni{\~n}o Southern Oscillation in a Changing Climate},
  pages = {65--86},
  address = {Washington DC},
  publisher = {AGU},
  year = {2020}
}

@article{hendon2007ensomjo,
  title={Seasonal dependence of the MJO--ENSO relationship},
  author={Hendon, Harry H and Wheeler, Matthew C and Zhang, Chidong},
  journal={Journal of climate},
  volume={20},
  number={3},
  pages={531--543},
  year={2007}
}

@article{hersbach2020era5,
  title={The ERA5 global reanalysis},
  author={Hersbach, Hans and Bell, Bill and Berrisford, Paul and Hirahara, Shoji and Hor{\'a}nyi, Andr{\'a}s and Mu{\~n}oz-Sabater, Joaqu{\'\i}n and Nicolas, Julien and Peubey, Carole and Radu, Raluca and Schepers, Dinand and others},
  journal={Quarterly journal of the royal meteorological society},
  volume={146},
  number={730},
  pages={1999--2049},
  year={2020},
  publisher={Wiley Online Library}
}

@article{huang2017ersstv5,
  title={Extended reconstructed sea surface temperature, version 5 (ERSSTv5): upgrades, validations, and intercomparisons},
  author={Huang, Boyin and Thorne, Peter W and Banzon, Viva F and Boyer, Tim and Chepurin, Gennady and Lawrimore, Jay H and Menne, Matthew J and Smith, Thomas M and Vose, Russell S and Zhang, Huai-Min},
  journal={Journal of Climate},
  volume={30},
  number={20},
  pages={8179--8205},
  year={2017}
}

@article{kamphuis2011retrograde,
  title={The global ocean circulation on a retrograde rotating earth},
  author={Kamphuis, V and Huisman, SE and Dijkstra, HA},
  journal={Climate of the Past},
  volume={7},
  number={2},
  pages={487--499},
  year={2011},
  publisher={Copernicus Publications G{\"o}ttingen, Germany}
}

@article{kikuchi2012bsiso,
  title={Bimodal representation of the tropical intraseasonal oscillation},
  author={Kikuchi, Kazuyoshi and Wang, Bin and Kajikawa, Yoshiyuki},
  journal={Climate Dynamics},
  volume={38},
  number={9},
  pages={1989--2000},
  year={2012},
  publisher={Springer}
}

@article{majda2009skeleton,
  title={The skeleton of tropical intraseasonal oscillations},
  author={Majda, Andrew J and Stechmann, Samuel N},
  journal={Proceedings of the National Academy of Sciences},
  volume={106},
  number={21},
  pages={8417--8422},
  year={2009},
  publisher={National Academy of Sciences}
}

@article{masunaga2007kwseaso,
  title={Seasonality and regionality of the Madden--Julian oscillation, Kelvin wave, and equatorial Rossby wave},
  author={Masunaga, Hirohiko},
  journal={Journal of the Atmospheric Sciences},
  volume={64},
  number={12},
  pages={4400--4416},
  year={2007}
}

@article{rasp2024weatherbench2,
  title={WeatherBench 2: A benchmark for the next generation of data-driven global weather models},
  author={Rasp, Stephan and Hoyer, Stephan and Merose, Alexander and Langmore, Ian and Battaglia, Peter and Russell, Tyler and Sanchez-Gonzalez, Alvaro and Yang, Vivian and Carver, Rob and Agrawal, Shreya and others},
  journal={Journal of Advances in Modeling Earth Systems},
  volume={16},
  number={6},
  pages={e2023MS004019},
  year={2024},
  publisher={Wiley Online Library}
}

@article{stachnik2015evaluating,
  title={Evaluating MJO event initiation and decay in the skeleton model using an RMM-like index},
  author={Stachnik, Justin P and Waliser, Duane E and Majda, Andrew J and Stechmann, Samuel N and Thual, Sulian},
  journal={Journal of Geophysical Research: Atmospheres},
  volume={120},
  number={22},
  pages={11--486},
  year={2015},
  publisher={Wiley Online Library}
}

@article{stechmann2015skeletondata,
  title={Identifying the skeleton of the Madden--Julian oscillation in observational data},
  author={Stechmann, Samuel N and Majda, Andrew J},
  journal={Monthly Weather Review},
  volume={143},
  number={1},
  pages={395--416},
  year={2015}
}

@article{thual2014stochastic,
  title={A stochastic skeleton model for the MJO},
  author={Thual, Sulian and Majda, Andrew J and Stechmann, Samuel N},
  journal={Journal of the Atmospheric Sciences},
  volume={71},
  number={2},
  pages={697--715},
  year={2014}
}

@article{wheelerkiladis1999,
  title={Convectively coupled equatorial waves: Analysis of clouds and temperature in the wavenumber--frequency domain},
  author={Wheeler, Matthew and Kiladis, George N},
  journal={Journal of the Atmospheric Sciences},
  volume={56},
  number={3},
  pages={374--399},
  year={1999}
}

@article{wheeler2004rmm,
  title={An all-season real-time multivariate MJO index: Development of an index for monitoring and prediction},
  author={Wheeler, Matthew C and Hendon, Harry H},
  journal={Monthly weather review},
  volume={132},
  number={8},
  pages={1917--1932},
  year={2004}
}

@article{zhang2005madden,
  title={Madden-julian oscillation},
  author={Zhang, Chidong},
  journal={Reviews of Geophysics},
  volume={43},
  number={2},
  year={2005},
  publisher={Wiley Online Library}
}

@article{zhang2020four,
  title={Four theories of the Madden-Julian oscillation},
  author={Zhang, C and Adames, {\'A}F and Khouider, B and Wang, B and Yang, D},
  journal={Reviews of Geophysics},
  volume={58},
  number={3},
  pages={e2019RG000685},
  year={2020},
  publisher={Wiley Online Library}
}


\clearpage

\pdfbookmark[1]{Tables}{section:sec_tables}
\section*{Tables}\label{sec_tables}

\begin{table}[h]
    \centering
    \begin{tabular}{ll}
        \toprule
        Conditioning & Description \\
        \midrule
        \texttt{pc1} & \begin{tabular}[t]{@{}l@{}}First principal component of the RMM-UBC index \\ (MJO phase/amplitude)\end{tabular} \\
        \texttt{pc2} & \begin{tabular}[t]{@{}l@{}}Second principal component of the RMM-UBC index \\ (MJO phase/amplitude)\end{tabular} \\
        \texttt{doyc} & Cosine component of the day-of-year (seasonal cycle) \\
        \texttt{doys} & Sine component of the day-of-year (seasonal cycle) \\
        \texttt{n34sst} & Niño 3.4 Sea Surface Temperature anomaly (ENSO state) \\
        \bottomrule
    \end{tabular}
    \caption{Low-dimensional conditioning variables used for climate prompting.}
    \label{tab:conditionings}
\end{table}

\begin{table}[h]
    \centering
    \begin{tabular}{lll}
        \toprule
        Model & Fields & Conditionings \\
        \midrule
        MJO Only& olr, ubc, zbc, q400 & pc1, pc2 \\
        MJO with Seasons& olr, ubc, zbc, q400 & pc1, pc2, doyc, doys \\
        MJO with Seasons/ENSO& olr, ubc & pc1, pc2, doyc, doys, n34sst \\
          \bottomrule
    \end{tabular}
    \caption{Trained Models.}
    \label{tab:trained_models}
\end{table}


\begin{table}[h]
    \centering
    \begin{tabular}{ll|ll}
        \toprule
        \multicolumn{4}{c}{\textbf{Model Architecture (U-Net)}} \\
        \midrule
        Levels (Hierarchy)   & 4              & Dim Multipliers      & (1, 2, 4, 8)        \\
        Bottleneck Size      & $2 \times 8$   & Attention Heads      & 8                   \\
        Positional Encoding  & Rotary (Temp)  & Sampling Factor      & $2\times$           \\
        \midrule
        \multicolumn{4}{c}{\textbf{Parameter (Training)}} \\
        \midrule
        Field width (lon)    & 64             & Field height (lat)   & 16                  \\
        Field Channels       & 2 to 4         & Field Frames (time)  & 16                  \\
        Conditionings        & 2 to 5         & Training steps       & 20,000              \\
        Condition dropout    & 0.1            & Dynamic threshold    & 99th quantile       \\
        \midrule
        \multicolumn{4}{c}{\textbf{Parameter (Sampling)}} \\
        \midrule
        Sampling method      & DDIM           & DDPM timesteps       & 1000                \\
        DDIM timesteps       & 250            & DDIM eta ($\eta$)    & 1                   \\
        Layering method      & Brick-wall     & Brick-wall stride    & 3                   \\
        Guidance ($w$)       & 1 (none)       &                      &                     \\
        \bottomrule
    \end{tabular}
    \caption{Model parameters for architecture, training, and sampling.}
    \label{tab:model_hyperparameters}
\end{table}


\clearpage


\clearpage

\pdfbookmark[1]{Figures}{section:sec_figures}
\section*{Figures}\label{sec_figures}

\clearpage

\clearpage

\clearpage

\clearpage

\clearpage

\clearpage

\clearpage

\end{appendices}
\end{document}